\documentclass[runningheads]{llncs}

\usepackage{eccv}

\usepackage{eccvabbrv}

\usepackage{graphicx}
\usepackage{booktabs}

\usepackage[accsupp]{axessibility}  

\usepackage{hyperref}

\usepackage{orcidlink}
\usepackage{booktabs}
\usepackage{multirow}

\begin{document}

\title{Beyond Natural-Image Foundation Models: Benchmarking Satellite Pretraining for Ophthalmic Image Analysis}

\titlerunning{Beyond Natural-Image Foundation Models}

\author{Lovre Antonio Budimir\thanks{
Corresponding author: \href{mailto:lovre-antonio.budimir@fer.unizg.hr}
{lovre-antonio.budimir@fer.unizg.hr}}\inst{,1,3}\orcidlink{0000-0003-2146-7067} \and
Mingya Alexa Gong\inst{2}\orcidlink{0009-0003-7337-875X} \and \\ Alyssa Foong Quinney\inst{3}\orcidlink{0009-0005-8692-1576} \and Ivana Matovinovi\'{c}\inst{1}\orcidlink{0000-0003-1498-7654} \and Yukun Zhou\inst{2,4,5,7}\orcidlink{0000-0002-0840-6422} \and \\ Pearse A. Keane\inst{2,5,6}\orcidlink{0000-0002-9239-745X} \and Sven Lon\v{c}ari\'{c}\inst{1}\orcidlink{0000-0002-4857-5351} \and Marinko V. \v{S}aruni\'{c}\inst{2,3,5}\orcidlink{0000-0002-5636-9056}}

\authorrunning{L.A.~Budimir \etal}


\institute{
Faculty of Electrical Engineering and Computing, University of Zagreb, HR \and
Institute of Ophthalmology, University College London, UK \and 
Medical Physics and Biomedical Engineering, University College London, UK \and
Department of Computer Science, University College London, UK \and
NIHR Moorfields Biomedical Research Centre, London, UK \and
Moorfields Eye Hospital NHS Foundation Trust, London, UK \and
Hawkes Institute, University College London, UK
}
\maketitle

\begin{figure}[t]
  \centering
  \includegraphics[width=\linewidth]{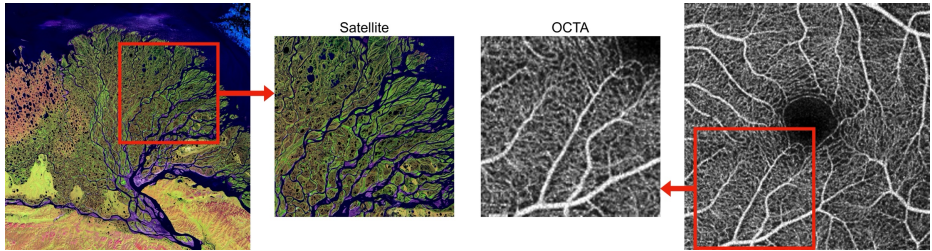}
  \caption{
   \textbf{Motivation.} Satellite image of the Lena River Delta in Russia (left; source: NASA, public domain) and an OCT angiography image (right; OCTA-500 dataset \cite{octa500}). Zoomed-in regions in the middle highlight the structural similarities between branching river networks and retinal vascular patterns observed in OCTA. 
  }
  \label{fig:motivation}
\end{figure}

\begin{abstract}

  Vision Foundation Models (VFMs) have emerged as a promising approach in medical imaging, producing broadly applicable systems that can be efficiently adapted across diverse imaging modalities, anatomical regions, and clinical tasks. However, VFMs require extensive training data, and their progress in medical image analysis is constrained by limited data availability, privacy concerns, and high development costs. To alleviate these constraints, medical VFMs (MedVFMs) are often built upon weights from generalist models pretrained on vast amounts of publicly available natural images, introducing a substantial distribution shift for medical task adaptation. To address this, we propose satellite imagery as a novel pretraining domain for MedVFM development and benchmarking, motivated by its closer visual alignment with medical data and its freedom from the privacy constraints that limit medical datasets. Across multiple ophthalmic imaging modalities, we compare DINOv3-SAT493m pretrained on 493 million satellite images against DINOv3-LVD1689m pretrained on 1.7 billion natural images, together with two medical specialist baselines: DINOv3-RETFound and MAE-RETFound. Our experiments show that satellite imagery is a stronger pretraining source than natural images for ophthalmic tasks, particularly on en face vascular-rich modalities. On several tasks, satellite pretraining matches or exceeds the medical specialists on high-resolution en face inputs, despite using no medical data. 

\end{abstract}

\section{Introduction}
\label{sec:introduction}

In recent years, the field of artificial intelligence (AI) has undergone a significant paradigm shift driven by the rapid development of foundation models (FMs) \cite{Bommasani2021FoundationModels,fm_in_med_imaging}. These models learn general-purpose feature representations from large-scale and diverse datasets, using self-supervised learning (SSL), and can be adapted to a broad range of downstream tasks with little or no task-specific fine-tuning \cite{dinov2,dinov3}. This paradigm holds substantial promise for medical fields, such as ophthalmology, where expert-annotated data are scarce and costly to obtain \cite{van2025foundation,retfound}. Diverse non-invasive imaging modalities provide complementary views of the retina and retinal vasculature that may support the assessment of both ocular conditions \cite{benchmark_ocular_disiase} and systemic health \cite{oculomics,oculomics_pk,oculomics_wagner}.

The success of FMs is largely enabled by access to vast amounts of data that can be easily obtained from the web. In computer vision, the \mbox{DINOv3~\cite{dinov3}} model has achieved remarkable performance across tasks such as image classification, semantic segmentation, and object detection, supported by training on approximately 1.689 billion natural images and scaling the Vision Transformer (ViT)~\cite{vit} encoder to roughly 7 billion parameters. However, developing medical vision foundation models (MedVFMs) at a comparable scale remains difficult due to several challenges. First, the acquisition of large-scale medical datasets is constrained by privacy, ethical, and regulatory concerns. Since medical data are difficult to share across institutions, resulting models may underperform for underrepresented populations, diseases, imaging devices, and clinical sites~\cite{ai_bias}. Second, developing MedVFMs requires substantial computational and financial resources, limiting such efforts primarily to well-resourced laboratories and medical institutions~\cite{retfound_green}.

These challenges are particularly relevant in ophthalmology, where clinicians rely on multiple non-invasive imaging modalities, including optical coherence tomography (OCT), OCT angiography (OCTA), color fundus photography (CFP), ultra-widefield imaging (UWF), and adaptive optics scanning laser ophthalmoscopy (AOSLO), for the diagnosis and monitoring of ocular and systemic diseases~\cite{visionFM}. 
To mitigate the cost of training from scratch, MedVFMs are often built on top of foundation models pretrained on natural images, followed by domain-specific continued pretraining or task-specific fine-tuning~\cite{generalist_vs_specialist}. 
\mbox{RETFound~\cite{retfound}}, a pioneering work in the field, was trained on 1.6 million OCT and CFP images using a masked autoencoder (MAE)~\cite{mae} already pretrained on 1.4 million natural images \cite{imagenet}. RETFound's successors, DINOv2-RETFound~\cite{dinov2_retfound} and RETFound-Green~\cite{retfound_green}, are built on top of DINOv2~\cite{dinov2}, which was pretrained on 142 million natural images. RETFound-DE~\cite{retfound_de} extends \cite{retfound} with additional pretraining on approximately 1 million synthetic CFP images.
Existing ophthalmic MedVFMs~\cite{retfound,dinov2_retfound,retfound_green, mirage} have primarily been evaluated on disease classification tasks using OCT and CFP images, and their development relies heavily on natural-image pretraining as the starting point.

The recent success and scale of DINOv3~\cite{dinov3} have raised the question of whether domain-specific additional pretraining on medical data remains necessary, or whether a generalist model trained on approximately 1.7 billion natural images can replace specialized MedVFMs. However, the substantial distribution shift between natural and ophthalmic images suggests that domain-specific models remain important for medical image analysis~\cite{dinov3_medstudy,generalist_vs_specialist}. In this work, we raise a complementary research question: \textit{what type of publicly available data, beyond natural images, can be more effectively leveraged for medical image analysis and MedVFM development?}

\begin{figure}[tb]
  \centering
  \includegraphics[width=\linewidth]{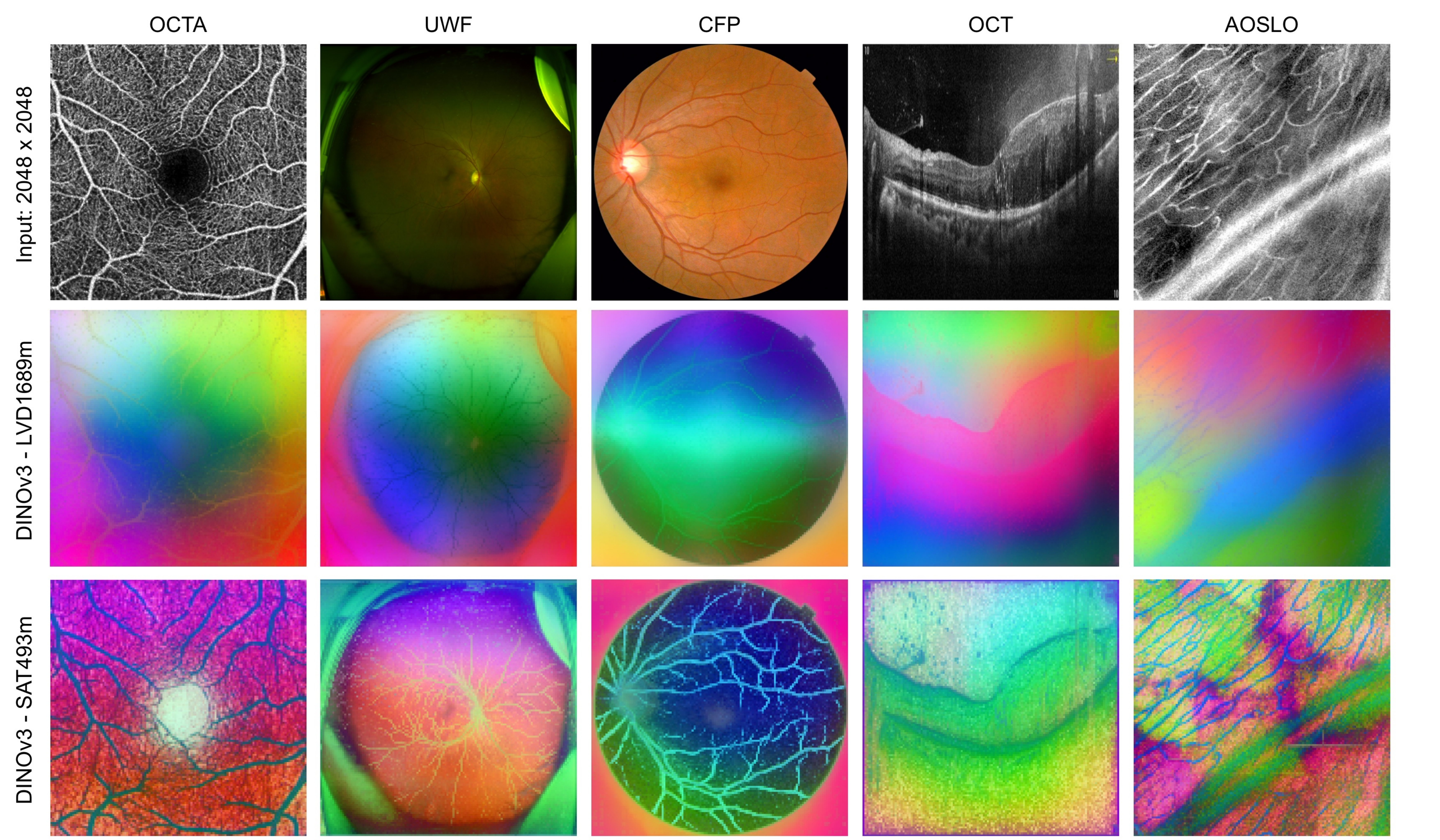}
  \caption{Visualization of dense features extracted from DINOv3-LVD1689m (middle) and DINOv3-SAT493m (bottom) across five ophthalmic imaging modalities: OCTA, UWF, CFP, OCT, and AOSLO. The first three principal components of a PCA computed over the feature space are mapped to RGB for visualization. 
  }
  \label{fig:pca_comparison}
\end{figure}

Natural image datasets typically consist of everyday objects and scenes, such as animals, vehicles, food, and people, which can be easily captured with mobile cameras. However, these data differ substantially in visual appearance from images acquired using ophthalmic imaging devices. In ophthalmology, MedVFMs must learn representations that capture fine-grained anatomical and pathological patterns, including retinal vessels, capillary networks, and subtle structural changes associated with disease. For this reason, we investigate satellite imagery as a potentially more aligned pretraining domain for ophthalmic representation learning. We are motivated by the connections between tubular structures in satellite images, such as roads and rivers, and vascular structures in retinal images, such as vessels and capillaries (see~\cref{fig:motivation}). Similar to natural images, satellite imagery can be acquired more easily and is subject to fewer privacy constraints compared to medical data, making it a more scalable source for building large-scale VFM datasets.

\begin{figure}[tb]
  \centering
  \includegraphics[width=0.9\linewidth]{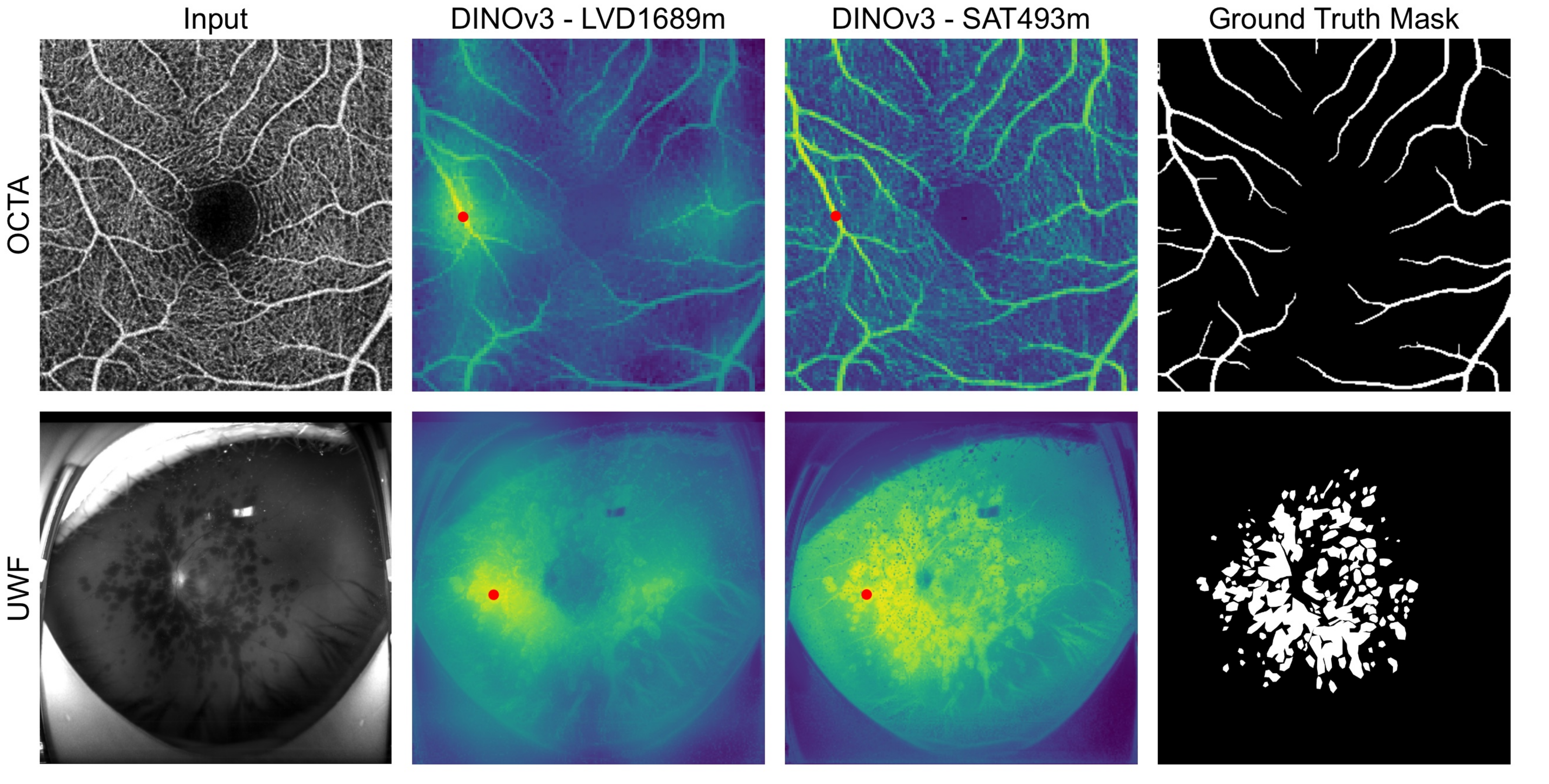}
  \caption{Cosine similarity scores computed between the dense feature at the red query patch and all other patch features for DINOv3-LVD1689m and DINOv3-SAT493m. DINOv3-SAT493m results in more consistent similarity scores along OCTA vessels (top) and UWF hemorrhagic regions (bottom), better matching the corresponding ground truth masks.
  }
  \label{fig:cosine_sim}
\end{figure}

To validate our hypothesis, we conduct a comprehensive benchmarking analysis using DINOv3~\cite{dinov3}, a state-of-the-art VFM that provides high-quality dense visual features. 
To evaluate representation quality on medical tasks, we compare frozen dense features extracted from two DINOv3 variants pretrained on different non-medical domains: DINOv3-LVD1689m, trained on 1.689 billion natural images, and DINOv3-SAT493m, trained on 493 million satellite images. We evaluate dense features from both models on nine ophthalmic datasets across five imaging modalities: OCTA, UWF, CFP, OCT, and AOSLO (see~\cref{fig:pca_comparison}). 
Our experiments show that satellite imagery provides a stronger pretraining source than natural images for segmentation on en face retinal scans (see \cref{fig:cosine_sim}). Additionally, we compare against DINOv3-RETFound~\cite{generalist_vs_specialist} and \mbox{MAE-RETFound}~\cite{retfound}, showing that \mbox{DINOv3-SAT493m}~\cite{dinov3} can outperform specialized MedVFMs for high-resolution en face inputs despite using no medical data during pretraining.   

This work studies how two publicly available non-medical pretraining domains, satellite imagery and natural images, align with medical downstream tasks. We evaluate both domains across five ophthalmic imaging modalities and a diverse set of diseases, anatomical structures, and pathological features. Our findings highlight satellite imagery as a novel, scalable, privacy-preserving, and structurally relevant pretraining source for the future development and deployment of medical vision foundation models.

\section{Related Work}
\label{sec:related_work}

\subsection{Medical Vision Foundation Models in Ophthalmology}

In ophthalmology, CFP and OCT are among the most widely used imaging modalities, making them the largest and most accessible sources of image data for developing MedVFMs~\cite{retfound}. RETFound~\cite{retfound} leverages this by developing two modality-specific models trained on approximately 904,170 CFP images and 736,442 OCT images, respectively. More recently, DINOv2-RETFound~\cite{dinov2_retfound} and \mbox{DINOv3-RETFound}~\cite{generalist_vs_specialist} use the same CFP training data, but replace the natural-image weights and MAE~\cite{mae} with DINOv2~\cite{dinov2} and DINOv3~\cite{dinov3}, respectively. RETFound-DE~\cite{retfound_de} addresses privacy constraints and the difficulty of acquiring large-scale medical datasets by training across three data domains. It follows the SSL objective of RETFound~\cite{retfound} and initializes its MAE backbone with ImageNet-1K pretrained weights, corresponding to approximately 1.4 million natural images. The model is then further pretrained on approximately 1 million synthetic CFP images, followed by final pretraining on 150 thousand real-world CFP images. MIRAGE~\cite{mirage} combines OCT and scanning laser ophthalmoscopy (SLO) data to develop a multi-modal MedVFM, while VisionFM~\cite{visionFM} considers a broader set of ophthalmic imaging modalities but still develops modality-specific MedVFMs. 
Existing work largely relies on natural image initialization, while the potential of structurally aligned, publicly available non-medical domains for learning ophthalmic representations remains underexplored.

\subsection{DINOv3 in Medical Imaging}

Motivated by the scale and strong performance of DINOv3~\cite{dinov3}, several recent works have adopted its features for medical vision tasks~\cite{meddinov3,generalist_vs_specialist,neuroseg,dinounet,beyond_natural}. To evaluate the quality of DINOv3 features in medical imaging, Liu \etal~\cite{dinov3_medstudy} benchmarked DINOv3 against MedVFMs across multiple imaging modalities and 2D/3D classification, 2D/3D segmentation, and registration tasks. In ophthalmology, Zhou \etal~\cite{generalist_vs_specialist} compared DINO-family models with MedVFMs pretrained on CFP images for classification. However, both studies primarily focus on the \mbox{DINOv3-LVD1689m} family of models pretrained on natural images, leaving it unclear how DINOv3 models pretrained on other large-scale domains, such as satellite imagery, transfer to medical imaging.

\begin{table*}[t]
\centering
\caption{Overview of retinal imaging datasets used in the benchmark setup.}
\label{tab:datasets}
\resizebox{\linewidth}{!}{
\begin{tabular}{lclcc}
\toprule
\textbf{Dataset} & \textbf{Modality} & \textbf{Task} & \textbf{\# Images} & \textbf{Resolution} \\
\midrule
 &  & \textbf{Segmentation} & &  \\

OCTA-500 (3M)~\cite{octa500} &
OCTA &
Large Vessels, Artery, Vein, FAZ &
200 &
$304 \times 304$ \\

OCTA-500 (6M)~\cite{octa500} &
OCTA &
Large Vessels, Artery, Vein, FAZ &
300 &
$400 \times 400$ \\

IDRiD~\cite{idrid} &
CFP &
4 lesion types, optic disc &
81 &
$4288 \times 2848$  \\

FIVES~\cite{fives} &
CFP &
Vessels&
800 &
$2048 \times 2048$  \\

UWF-RHS~\cite{uwf_rhs} &
UWF &
Hemorrhages &
215 &
$3900 \times 3072$  \\ 

PRIME-FP20~\cite{prime_fp20} &
UWF &
Vessels &
15 &
$4000 \times 4000$  \\

AMD-SD~\cite{hu2024amd} &
OCT &
5 lesion types &
3049 &
$570 \times 380$  \\

HCMS~\cite{he2018retinal}  &
OCT &
8 retinal layers &
1715 &
$1024 \times 496$ \\

AOVesselCNN~\cite{aovesselcnn} &
AOSLO &
Capillaries &
199 &
$768 \times 768$  \\

\midrule
 &  & \textbf{Classification} & &  \\

MMRDR-CFP~\cite{mmrdr} &
CFP &
DR grading (5 classes) &
11,118 &
Various \\

MMRDR-UWF~\cite{mmrdr} &
UWF &
DR grading (5 classes) &
10,404 &
Various \\

MMRDR-OCT~\cite{mmrdr} &
OCT &
DME grading (3 classes) &
2,938 &
Various  \\

\bottomrule
\end{tabular}}
\end{table*}
\section{Datasets}
\label{sec:datasets}

We benchmark models on nine publicly available ophthalmic datasets across five imaging modalities: OCTA, CFP, UWF, OCT, and AOSLO. The selected datasets cover diverse image resolutions, dataset sizes, diseases, and anatomical and pathological structures. They also cover a wide range of fields of view (FOV), from $2^{\circ}$ in AOSLO images~\cite{aovesselcnn} to $200^{\circ}$ in UWF images~\cite{prime_fp20}. A dataset overview is provided in \cref{tab:datasets}, with additional dataset-specific details described in the remainder of this section.

\subsection{OCTA Segmentation Dataset}

\subsubsection{OCTA-500~\cite{octa500}} is a public OCTA dataset collected from 500 subjects using two different FOVs. Following the original dataset split, we denote the $6~\mathrm{mm} \times 6~\mathrm{mm}$ FOV subset with 300 subjects as \textbf{OCTA-500 (6M)}, and the $3~\mathrm{mm} \times 3~\mathrm{mm}$ FOV subset with 200 subjects as \textbf{OCTA-500 (3M)}. The two subsets contain both normal subjects and subjects with different ophthalmic diseases, including age-related macular degeneration (AMD), diabetic retinopathy (DR), and choroidal neovascularization (CNV). In our benchmark setup, model inputs are OCTA maximum projection maps generated between the internal limiting membrane (ILM) and outer plexiform layer (OPL), while the ground-truth annotations consist of four segmentation masks: large vessel, artery, vein, and foveal avascular zone (FAZ). In both subsets, we follow \cite{octa500} and use 50 subjects for testing.

\subsection{CFP Segmentation Datasets}

\subsubsection{IDRiD~\cite{idrid}} is a public CFP dataset released as part of the ``Diabetic Retinopathy: Segmentation and Grading Challenge.'' 
In our benchmark setup, we use 81 CFP images from the segmentation sub-challenge for typical diabetic retinopathy lesions and optic disc segmentation. We follow the official training/test split from~\cite{idrid} and evaluate performance on segmentation of the optic disc and four lesion types: hemorrhages, microaneurysms, hard exudates, and soft exudates.

\subsubsection{FIVES~\cite{fives}} is a public CFP dataset for retinal vessel segmentation. The dataset consists of 600 training images and 200 testing images, with pixel-level annotations of retinal vessels. It includes images from healthy subjects as well as subjects with ophthalmic diseases, including glaucoma, AMD, and DR.

\subsection{UWF Segmentation Datasets}

\subsubsection{UWF-RHS~\cite{uwf_rhs}} is a public UWF retinal hemorrhage segmentation dataset collected from 215 subjects.
The dataset primarily includes patients with DR, along with a smaller proportion of patients with other ophthalmic diseases, such as AMD and retinal vein occlusion (RVO). In our benchmark setup, we randomly split the dataset into training and test sets using a 70/30 ratio for semantic segmentation evaluation.

\subsubsection{PRIME-FP20~\cite{prime_fp20}} is a public ultra-widefield (UWF) fundus dataset for retinal vessel segmentation. It consists of 15 high-resolution UWF fundus photographs.
In our benchmark setup, we randomly split the dataset into training and test sets using a 70/30 ratio for vessel segmentation evaluation. We additionally pre-process the images by cropping black background regions outside the FOV.

\subsection{OCT Segmentation Datasets}
\subsubsection{HCMS~\cite{he2018retinal}} is a public OCT dataset for retinal layer segmentation. It contains images of 14 healthy subjects and 21 subjects with diagnosed multiple sclerosis (MS). For each subject, there are 49 OCT B-scans with 8 labelled retinal layers: retinal nerve fibre layer (RNFL),
ganglion cell layer and inner plexiform layer (GCL+IPL),
inner nuclear layer	(INL), outer plexiform layer (OPL), outer nuclear layer (ONL), inner photoreceptor segment (IS), outer photoreceptor segments (OS), and retinal pigment epithelium. We include 735 images for the training phase and evaluate the models on 980 images.

\subsubsection{AMD-SD~\cite{hu2024amd}} is a public OCT retinal dataset for wet AMD lesion segmentation. It comprises 3049 OCT B-scan images from 138 subjects with five segmentation labels: intraretinal fluid (IRF), subretinal fluid (SRF), pigment epithelial detachment (PED), subretinal hyperreflective material (SHRM), and IS/OS junction disruption. We follow the splitting protocol from~\cite{hu2024amd},  with  2346 images in the training set and 703 images in the test set.

\subsection{AOSLO Segmentation Dataset}

AOSLO is a high-resolution retinal imaging modality primarily used in research settings, enabling visualization of microscopic retinal structures and cells, such as photoreceptors and white blood cells \cite{aoslo_info}. In our benchmark setup, we use the public dataset from~\cite{aovesselcnn} for capillary segmentation. The dataset contains 199 manually annotated images derived from 27 AOSLO perfusion montages, including data from five healthy human eyes, six healthy non-human primate eyes, and five non-human primate eyes with laser-induced glaucoma~\cite{aovesselcnn}. Following~\cite{aovesselcnn}, we use 185 images for training and 14 images for testing.

\subsection{Classification on CFP, UWF, and OCT}

\textbf{MMRDR~\cite{mmrdr}} is a public multi-modal retinal image dataset for DR and DME grading and lesion classification, containing three imaging modalities: UWF imaging, CFP, and OCT. Each modality is provided as a separate subset, consisting of 8,893 training and 2,225 test CFP images, 7,807 training and 2,597 test UWF images, and 2,376 training and 562 test OCT images. In our benchmark setup, we evaluate CFP and UWF images using multi-class classification with five DR severity grades, while OCT images are evaluated using three diabetic macular edema (DME) severity grades.

\section{Benchmark Setup}
\label{sec:setup}

\subsection{Models}
\label{sec:models}

We evaluate on all datasets using two non-medical models from DINOv3~\cite{dinov3}: \mbox{DINOv3-LVD1689m} and \mbox{DINOv3-SAT493m}. \mbox{DINOv3-LVD1689m} is trained on 1.689 billion natural images, whereas \mbox{DINOv3-SAT493m} is trained on 493 million satellite images\cite{dinov3}. For CFP datasets, we include \mbox{DINOv3-RETFound}~\cite{generalist_vs_specialist}, a domain-specific model initialized from \mbox{DINOv3-LVD1689m} and further trained on 904 thousand CFP images from Moorfields Eye Hospital~\cite{generalist_vs_specialist,dinov2_retfound}. We also evaluate \mbox{MAE-RETFound}~\cite{retfound,dinov2_retfound}, another domain-specific model, on CFP and OCT images. To ensure a fair comparison between pretraining domains, all models use the same ViT-Large architecture~\cite{vit}.

\subsection{Task Adaptation}
\label{sec:adaptation}

The DINOv3 model family produces high-quality dense visual features and has shown strong performance on dense computer vision tasks, such as semantic segmentation, without additional fine-tuning~\cite{dinov3}. To evaluate dense feature quality across multiple pretraining domains on ophthalmic datasets, we freeze the VFM backbone and train a lightweight head module for all tasks.

For segmentation tasks, VFM inputs are resized to $1024 \times 1024$, except for PRIME-FP20~\cite{prime_fp20}, where inputs are resized to $4096 \times 4096$ due to the small dataset size. All inputs are normalized, and no additional data augmentation is applied. The linear head is trained for 30 epochs using the AdamW optimizer~\cite{adamw} with a cosine learning rate scheduler, an initial learning rate of $0.01$, and a weight decay of $10^{-5}$. We use a batch size of 8 and optimize the model with Dice loss, except for OCT datasets, where we use cross-entropy loss. The predicted masks are up-sampled to $512 \times 512$ resolution, and binary segmentation masks are obtained using a threshold of $0.5$. 

For classification tasks, we train an attention-based multiple instance learning (MIL) module~\cite{mil} on top of the frozen dense features. Since classification does not require pixel-level predictions and is evaluated on larger datasets, inputs are resized to $512 \times 512$ and normalized for all tasks. The MIL module uses a hidden dimension of 256 and is trained for 30 epochs with AdamW~\cite{adamw}, a cosine learning rate scheduler, an initial learning rate of $5 \times 10^{-4}$, and a weight decay of $10^{-5}$. We use a five-epoch warm-up and weighted cross-entropy loss during training. 

All results are reported on the test set using the final training checkpoint. We report mean values over five and three runs for segmentation and classification tasks, respectively. Experiments are implemented in PyTorch and performed on an NVIDIA RTX 4090 GPU.

\subsection{Metrics}
\label{sec:metrics_implementation}

We report Dice score for segmentation tasks, and accuracy (Acc), quadratic weighted kappa (QWK), and weighted F1 score (wF1) for classification tasks.

\section{Results}
\label{sec:results}

\begin{table}[t]
\centering
\caption{Quantitative comparison on OCTA segmentation task using OCTA-500~\cite{octa500} dataset. Results are reported as Dice scores (\%). The best result is shown in bold.}
\label{tab:octa_results}
\begin{tabular}{lcccccccc}
\toprule
& \multicolumn{4}{c}{OCTA-500 (3M)~\cite{octa500}} & \multicolumn{4}{c}{OCTA-500 (6M)~\cite{octa500}} \\
\cmidrule(lr){2-5} \cmidrule(lr){6-9}
\multicolumn{1}{c}{\multirow{-2}{*}{OCTA}} & Vessel & Artery & Vein & FAZ & Vessel & Artery & Vein & FAZ \\
\midrule
DINOv3-LVD1689m~\cite{dinov3} & 68.2 & 58.3 & 54.7 & \textbf{94.7} & 61.6 & 47.8 & 51.8 & 86.1  \\
DINOv3-SAT493m~\cite{dinov3}  & \textbf{71.8} & \textbf{62.7} & \textbf{60.4} & 94.1 & \textbf{64.4} & \textbf{52.1} & \textbf{55.3 }& \textbf{87.9} \\
\bottomrule
\end{tabular}
\end{table}

\subsection{OCTA Segmentation Results}
\label{sec:octa_results}

\Cref{tab:octa_results} presents results on OCTA-500~\cite{octa500}. DINOv3-SAT493m~\cite{dinov3}, pretrained on satellite imagery, outperforms DINOv3-LVD1689m, pretrained on natural images, on all vascular segmentation tasks, including arteries, veins, and their union, \ie, the large vessel task. For FAZ segmentation, results are closer and  inconsistent, with the DINOv3-LVD1689m model achieving slightly better performance with the smaller FOV, where the task appears less challenging.

\begin{table}[htbp]
\centering
\caption{Quantitative comparison on CFP segmentation task. Left: performance on five segmentation tasks from the IDRiD dataset~\cite{idrid}, covering diabetic retinopathy lesions and retinal structures: hemorrhages (HE), microaneurysms (MA), hard exudates (EX), soft exudates (SE), and optic disc (OD). Right: performance on retinal vessel segmentation from the FIVES dataset~\cite{fives}. Results are reported as Dice scores (\%). The best result is shown in bold and the second-best result is underlined.}
\label{tab:idrid_fives}
\setlength{\tabcolsep}{4pt}
\begin{tabular}{lccccccc}
\toprule
& \multicolumn{6}{c}{IDRiD~\cite{idrid}} & \multicolumn{1}{c}{FIVES~\cite{fives}} \\
\cmidrule(lr){2-7} \cmidrule(lr){8-8}
\multicolumn{1}{c}{\multirow{-2}{*}{CFP}} & HE & MA & EX & SE & OD & Overall & Vessel \\
\midrule
MAE-RETFound~\cite{retfound} & 32.4 & \underline{10.2} & \textbf{41.9} & 37.0 & \underline{94.5}  & 43.2 & 57.0 \\
DINOv3-RETFound~\cite{generalist_vs_specialist} & 32.6 & 8.7 & 40.4 & 36.0 & 94.2  & 42.4 & 54.6 \\
\midrule
DINOv3-LVD1689m~\cite{dinov3}  & \textbf{38.1} & 8.4 & 40.4 & \underline{46.9} & \textbf{94.7}  & \underline{45.7} & \textbf{60.6} \\
DINOv3-SAT493m~\cite{dinov3}  & \underline{37.0} & \textbf{13.2} & \textbf{41.9} & \textbf{54.5} & 93.5  & \textbf{48.0} & \underline{60.5} \\
\bottomrule
\end{tabular}
\end{table}

\subsection{CFP Segmentation Results}
\label{sec:cfp_results}

\Cref{tab:idrid_fives} presents results on the IDRiD~\cite{idrid} dataset and the FIVES~\cite{fives} dataset. On IDRiD, DINOv3-SAT493m overall achieves the best performance on retinopathy lesion segmentation, outperforming \mbox{DINOv3-LVD1689m} and the two CFP domain-specific models, DINOv3-RETFound~\cite{generalist_vs_specialist} and \mbox{MAE-RETFound}~\cite{retfound}. In contrast, DINOv3-LVD1689m performs better on optic disc segmentation, which appears to be a simpler anatomical segmentation task. On FIVES, the two non-medical models show comparable vessel segmentation performance and outperform the domain-specific models.

\begin{table}[t]
\centering
\caption{Quantitative comparison on UWF segmentation task. Left: performance on retinal hemorrhage segmentation from the UWF-RHS dataset~\cite{uwf_rhs}. Right: performance on retinal vessel segmentation from the PRIME-FP20 dataset~\cite{prime_fp20}. Results are reported as Dice scores (\%) and the best is shown in bold.}\label{tab:uwf}
\setlength{\tabcolsep}{4pt}
\begin{tabular}{lcc}
\toprule
& \multicolumn{1}{c}{UWF-RHS~\cite{uwf_rhs}} & \multicolumn{1}{c}{PRIME-FP20~\cite{prime_fp20}} \\

\cmidrule(lr){2-2} \cmidrule(lr){3-3}
\multicolumn{1}{c}{\multirow{-2}{*}{UWF}} & Hemorrhages & Vessel \\
\midrule
DINOv3-LVD1689m~\cite{dinov3} & 37.3 & 46.4 \\
DINOv3-SAT493m~\cite{dinov3}  & \textbf{42.5} & \textbf{47.9} \\
\bottomrule
\end{tabular}
\end{table}

\subsection{UWF Segmentation Results}
\label{sec:uwf_results}

\Cref{tab:uwf} presents results on the UWF-RHS~\cite{uwf_rhs} (left) and the PRIME-FP20~\cite{prime_fp20} (right) datasets. Across both UWF segmentation tasks, hemorrhage segmentation and vessel segmentation, DINOv3-SAT493m, pretrained on satellite imagery, outperforms DINOv3-LVD1689m, pretrained on natural images.

\begin{table}[t]
\centering
\caption{Quantitative comparison on OCT HCMS~\cite{he2018retinal} retinal layer segmentation and AMD-SD~\cite{hu2024amd} lesion segmentation tasks. Results are reported as Dice scores (\%). The best is shown in bold.}
\label{tab:oct_seg}
\resizebox{\linewidth}{!}{
\setlength{\tabcolsep}{4pt}
\begin{tabular}{lcccccccc ccccc}
\toprule
\multirow{3}{*}{OCT}
& \multicolumn{8}{c}{HCMS~\cite{he2018retinal}}
& \multicolumn{5}{c}{AMD-SD~\cite{hu2024amd}} \\
\cmidrule(lr){2-9} \cmidrule(lr){10-14}
& \multicolumn{8}{c}{Retinal Layer}
& \multicolumn{5}{c}{Lesions} \\
\cmidrule(lr){2-9} \cmidrule(lr){10-14}
& RNFL & GCL+IPL & INL & OPL & ONL & IS & OS & RPE
& SRF & IRF & PED & SHRM & IS/OS \\
\midrule
DINOv3-LVD1689m~\cite{dinov3}
& 85.3 & 85.7 & 64.4 & 72.3 & 88.3 & 71.5 & 74.6 & 83.3
& 37.6 & \textbf{28.1} & \textbf{28.9} & \textbf{32.7} & 46.7 \\
DINOv3-SAT493m~\cite{dinov3}
& \textbf{87.6} & \textbf{88.8} & \textbf{71.2} & \textbf{77.2} & \textbf{90.4} & \textbf{77.2} & \textbf{80.3} & \textbf{86.0}
& \textbf{40.3} & 24.7 & 28.2 & 32.5 & \textbf{48.9} \\
\bottomrule
\end{tabular}}
\end{table}
\subsection{OCT Segmentation Results}
\label{sec:oct_results}
\Cref{tab:oct_seg} (left) presents results on the OCT segmentation task of 8 retinal layers \cite{he2018retinal}. \mbox{DINOv3-SAT493m}~\cite{dinov3} outperforms \mbox{DINOv3-LVD1689m}~\cite{dinov3}, achieving higher Dice scores across all 8 retinal layers. \Cref{tab:oct_seg} (right) presents results on the OCT segmentation task of 5 wet AMD lesions \cite{hu2024amd}. \mbox{DINOv3-LVD1689m}~\cite{dinov3} shows better performance on 3 out of 5 lesions, achieving a notable +3.4\% improvement for IRF.

\begin{table}[t]
\centering
\caption{Quantitative comparison on AOSLO capillary segmentation task. Results are reported as Dice scores (\%) and the best is shown in bold.}\label{tab:aoslo}
\setlength{\tabcolsep}{5pt}
\begin{tabular}{lc}
\toprule
AOSLO & Capillaries \\
\midrule
DINOv3-LVD1689m~\cite{dinov3} & 56.1  \\
DINOv3-SAT493m~\cite{dinov3}  & \textbf{60.2}  \\
\bottomrule
\end{tabular}
\end{table}
\subsection{AOSLO Segmentation Results}
\label{sec:aoslo_results}

\Cref{tab:aoslo} presents results on the AOSLO capillary segmentation task \cite{aovesselcnn}. The \mbox{DINOv3-SAT493m}~\cite{dinov3} model again outperforms \mbox{DINOv3-LVD1689m}~\cite{dinov3} on vascular structure segmentation, achieving a +4.1\% improvement in Dice score.

\begin{figure}[t]
  \centering
  \includegraphics[width=\linewidth]{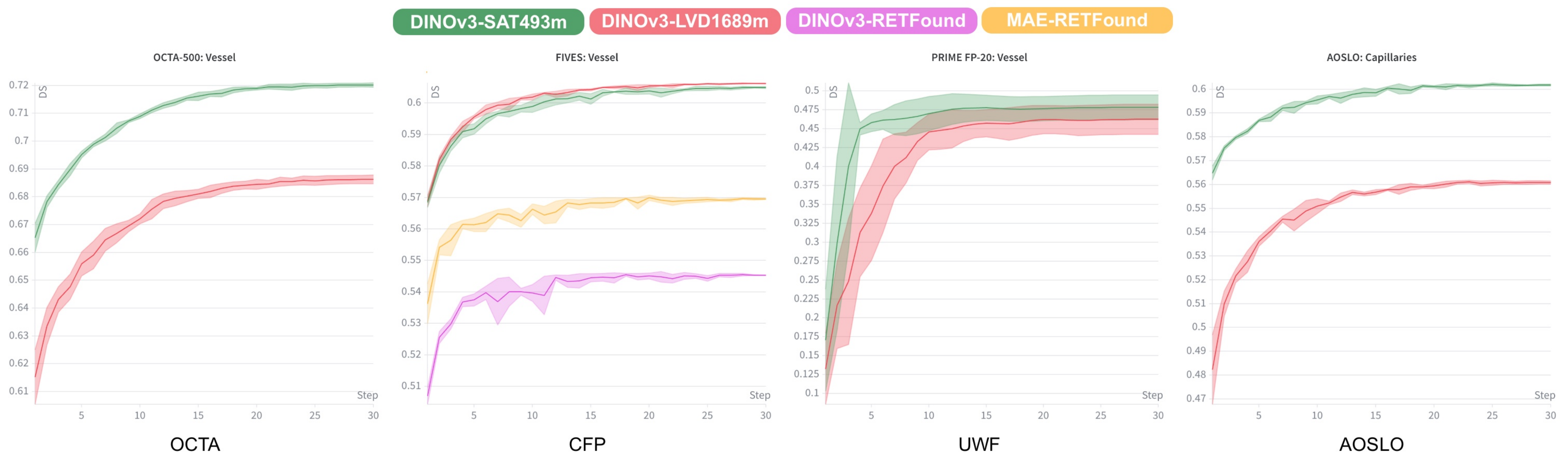}
      \caption{Dice score (DS) over training epochs for vasculature segmentation (vessels and capillaries) across four en face modalities:  OCTA, CFP, UWF, and AOSLO.
  }
  \label{fig:training}
\end{figure}

\begin{table*}[t]
\centering
\caption{Classification performance across retinal imaging modalities on MMRDR~\cite{mmrdr} dataset. Results are reported
as accuracy (Acc), quadratic weighted kappa (QWK), and weighted F1 score (wF1). Best results are highlighted in bold.}
\label{tab:classification_results}
\small
\setlength{\tabcolsep}{4pt}
\resizebox{\linewidth}{!}{
\begin{tabular}{lccccccccc}
\toprule

& \multicolumn{3}{c}{\textbf{CFP}}
& \multicolumn{3}{c}{\textbf{UWF}}
& \multicolumn{3}{c}{\textbf{OCT}} \\

\cmidrule(lr){2-4}
\cmidrule(lr){5-7}
\cmidrule(lr){8-10}

\textbf{Method}
& Acc & QWK & wF1
& Acc & QWK & wF1
& Acc & QWK & wF1 \\

\midrule

MAE-RETFound~\cite{retfound}
& 75.3 & 85.0 & 81.0
& -- & -- & --
& \textbf{88.9} & \textbf{84.7} & \textbf{88.8} \\

DINOv3-RETFound~\cite{generalist_vs_specialist}
& 74.3 & 82.5 & 73.7
& -- & -- & --
& -- & -- & -- \\
\midrule
DINOv3-LVD1689m~\cite{dinov3}
& 78.0 & 88.4 & 78.7
& 66.9 & 87.8 & 66.8
& 86.2 & 84.2 & 86.5 \\

DINOv3-SAT493m~\cite{dinov3}
& \textbf{80.6} & \textbf{89.5} & \textbf{81.0}
& \textbf{70.0} & \textbf{89.0} & \textbf{69.5}
& 83.9 & 81.8 & 85.0 \\

\bottomrule
\end{tabular}}
\end{table*}
\subsection{Classification Results}
\label{sec:cls_results}

\Cref{tab:classification_results} presents results on the multi-modal MMRDR~\cite{mmrdr} dataset across three imaging modalities: CFP, UWF, and OCT. On CFP, DINOv3-SAT493m~\cite{dinov3} outperforms DINOv3-LVD1689m~\cite{dinov3} as well as the two domain-specialized models, MAE-RETFound~\cite{retfound} and DINOv3-RETFound~\cite{generalist_vs_specialist}. On UWF, the satellite-pretrained VFM again outperforms the natural-image  VFM across all metrics. In contrast, on OCT, the domain-specific MAE-RETFound~\cite{retfound} model pretrained on OCT images outperforms both non-medical models, while DINOv3-SAT493m shows the worst overall performance.

\begin{figure}[!htb]
  \centering
  \includegraphics[width=\linewidth]{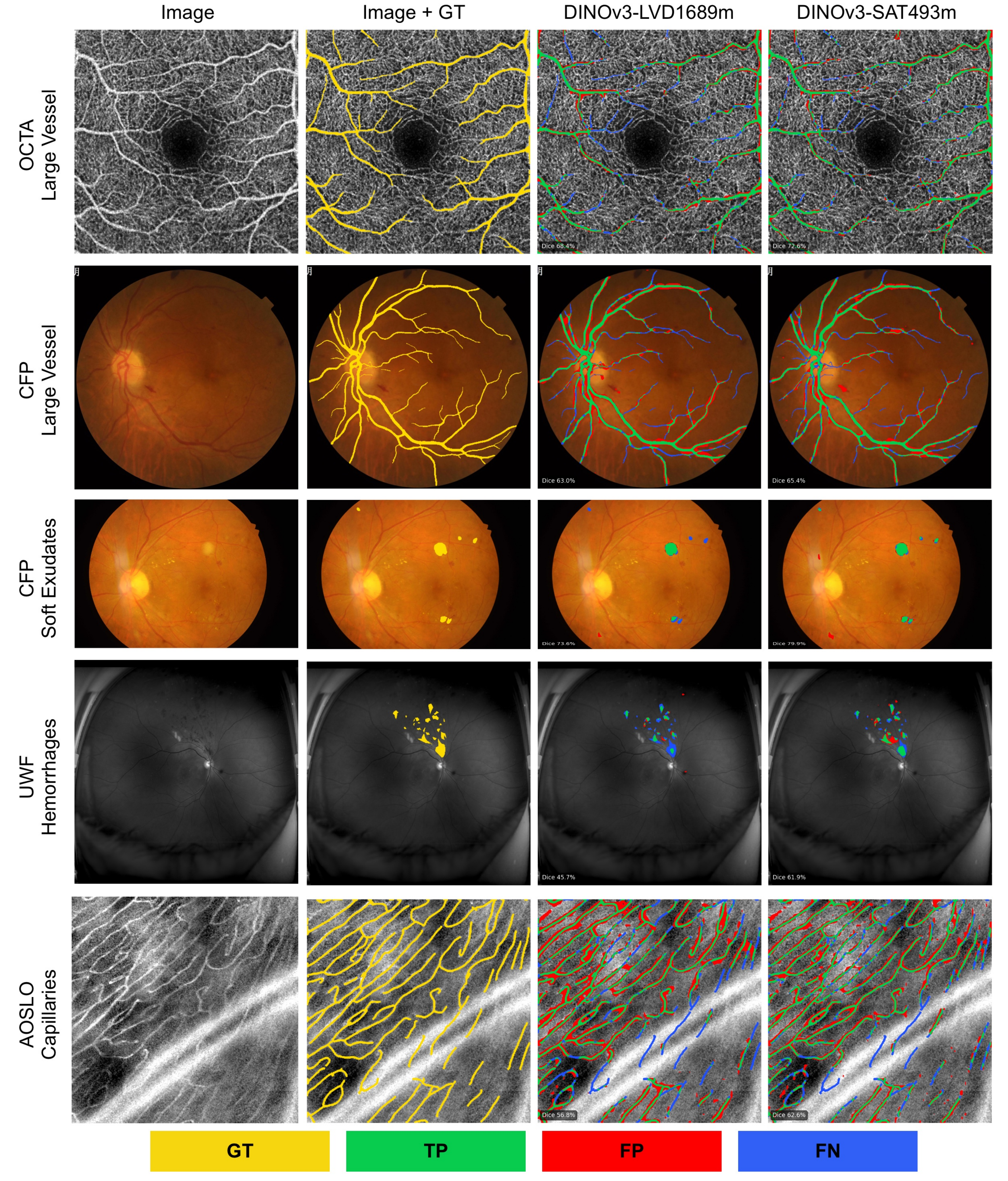}
  \caption{
   Qualitative results on segmentation tasks across four different modalities. 
  }
  \label{fig:qualitative}
\end{figure}

\begin{figure}[tb]
  \centering
  \includegraphics[width=\linewidth]{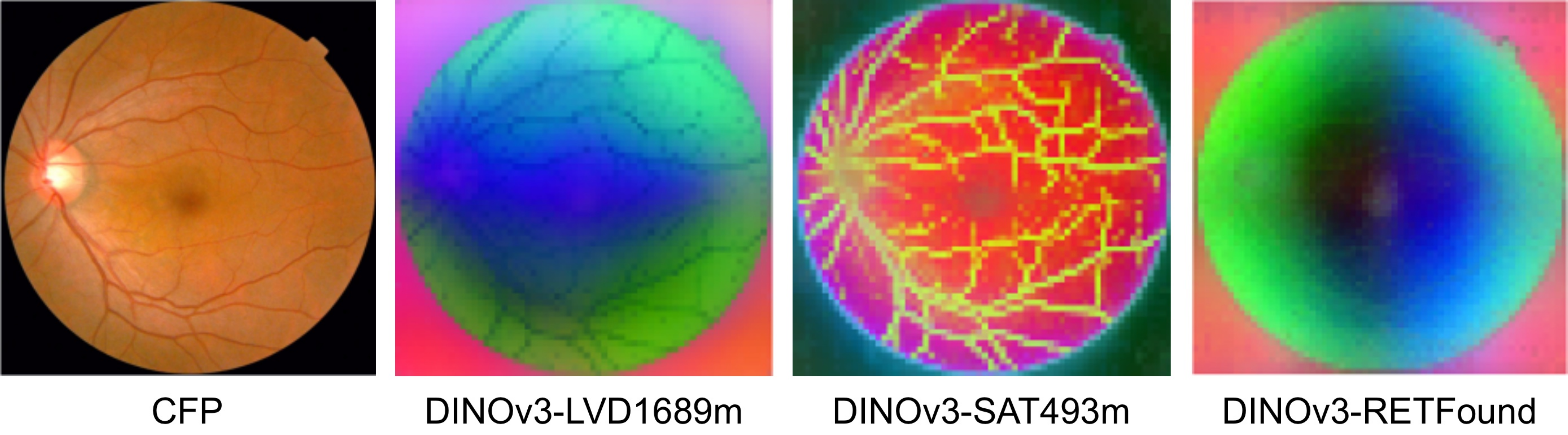}
  \caption{Visualization of dense CFP features extracted from DINOv3-LVD1689m, \mbox{DINOv3-SAT493m}, and DINOv3-RETFound. The first three principal components of a PCA computed over the feature space are mapped to RGB for visualization. Input image size to the DINOv3 model is $1024 \times 1024$.
  }
  \label{fig:pca_comparison_discussion}
\end{figure}

\section{Discussion}
\label{sec:discussion}

In this study, we closely evaluated frozen dense features from two non-medical DINOv3~\cite{dinov3} models, pretrained on natural (DINOv3-LVD1689m) and satellite images (DINOv3-SAT493m), across a diverse set of ophthalmic datasets. The datasets span multiple imaging modalities with substantially different fields of view, image resolutions, diseases, and anatomical and pathological structures. By benchmarking these two models, we analyze how different non-medical pretraining domains align with ophthalmic images and compare their downstream adaptation efficiency with each other and with domain-specialized medical foundation models (DINOv3-RETFound~\cite{generalist_vs_specialist} and MAE-RETFound~\cite{retfound}). These findings provide practical insights on selecting and leveraging non-medical pretraining data for medical imaging tasks, particularly when medical datasets are difficult to acquire. They also inform future efforts to develop and deploy clinically useful medical foundation models.

DINOv3-SAT493m~\cite{dinov3}, pretrained on satellite imagery, consistently outperforms the natural-image DINOv3-LVD1689m~\cite{dinov3} model across most segmentation and classification tasks. This difference is especially notable for en face imaging modalities: UWF, CFP, OCTA, and AOSLO. These modalities contain rich retinal vascular patterns, suggesting that DINOv3-SAT493m may be better suited for capturing tubular structures, such as vessels and capillaries, and preserving their spatial continuity across retinal images. This advantage is reflected in overall faster convergence during training and better performance on vessel and capillary segmentation (see \cref{fig:training}). In addition, \mbox{DINOv3-SAT493m} appears better suited for capturing fine-grained and spatially distributed structures, as reflected by its overall stronger performance on small pathological lesions such as soft exudates and hemorrhages (see~\cref{fig:qualitative}). Both effects may be partly explained by satellite-image pretraining, where thin elongated structures with sharp semantic boundaries (\eg, rivers and roads) and small spatially distributed objects (\eg, trees and houses) are common, potentially providing useful inductive biases for medical imaging.

Both non-medical models use the same ViT-Large architecture with approximately 300 million parameters, allowing a direct comparison of the effect of pretraining data. Despite being pretrained on more than one billion fewer images than \mbox{DINOv3-LVD1689m}, DINOv3-SAT493m~\cite{dinov3} achieves better overall performance across most tasks. This suggests that domain alignment may be more important than dataset scale alone when transferring vision foundation models to medical imaging tasks.

When comparing DINOv3-SAT493m and DINOv3-LVD1689m against the \mbox{DINOv3-RETFound}~\cite{generalist_vs_specialist} model on CFP images, we observe that both non-medical models outperform the domain-specific model on segmentation and classification tasks using dense features. In \cref{fig:pca_comparison_discussion}, we visualize dense features from all three models. The PCA representation of DINOv3-RETFound appears more similar to DINOv3-LVD1689m representations, which may be consistent with the fact that DINOv3-RETFound was initialized with this model. 
Additionally, DINOv3-RETFound was primarily adapted for downstream classification tasks and did not use the Gram anchoring loss, nor did it undergo the high-resolution refinement stage employed during the final stages of \mbox{DINOv3} training~\cite{dinov3}. As a result, its dense representations may be less optimized for high-resolution retinal image analysis, potentially explaining its weaker performance in our evaluation.

Despite our extensive evaluation, this work has a few limitations. First, our analysis is limited to ophthalmic imaging tasks and the DINOv3 ViT-Large architecture. Future work should extend this evaluation to other medical domains, VFM architectures, and pretraining procedures. Second, we evaluate segmentation tasks using linear probing, which may limit performance due to the spatial downsampling of ViT \cite{vit}. This limitation is particularly relevant for small lesions and anatomical structures that may be thinner or smaller than the model patch size. Future work should explore more segmentation and classification heads, as well as fine-tuning-based evaluation protocols.

\section{Conclusion}

In summary, we benchmark DINOv3~\cite{dinov3} using two scalable non-medical pretraining domains, satellite imagery and natural images, on ophthalmic downstream tasks. Our findings show that pretraining VFMs on satellite images results in more domain-aligned feature representations across multiple ophthalmic imaging modalities, positioning satellite imagery as a scalable, privacy-preserving, and structurally relevant source for medical image analysis and the future development of medical vision foundation models.

\section*{Acknowledgements}
L.A.~Budimir acknowledges support from the British Scholarship Trust (BST) and the Agency for Mobility and EU Programmes (AMPEU) for his research visit to University College London (UCL), during which he contributed to this project.
M.A.~Gong is supported by a BRC Translational Studentship (Imaging), funded by the Moorfields Biomedical Research Centre (BRC).
A.F.~Quinney is supported by the Engineering and Physical Sciences Research Council’s Co-operative Awards in Science and Engineering [EP/W524335/1], with industrial sponsorship provided by Optos Plc.
Y.~Zhou is supported by a Wellcome Trust Early-Career Award (318987/Z/24/Z) and a Moorfields Eye Charity Equipment Grant (EQR-25B-102).
P.A.~Keane is supported by a UK Research {\&} Innovation Future Leaders Fellowship (MR/T019050/1), Moorfields Eye Charity with The Rubin Foundation Charitable Trust (GR001753), and an Alcon Research Institute Senior Investigator Award.
S.~Lončarić and I.~Matovinović are supported by the European Regional Development Fund under grant agreement PK.1.1.10.0007 (DATACROSS).
M.V.~Šarunić is supported by Moorfields Eye Charity and the NIHR Biomedical Research Centre at Moorfields Eye Hospital and UCL Institute of Ophthalmology.


%
%
\bibliographystyle{splncs04}
\bibliography{main}
\end{document}